\documentclass[preprint,12pt]{elsarticle}

\usepackage{amssymb}
\usepackage{graphicx}
\usepackage{amsmath}
\usepackage{subfigure}
\usepackage{float}
\usepackage{algorithm}
\usepackage{algorithmic}
\usepackage{multirow}

\journal{Nuclear Physics B}

\begin{document}

\begin{frontmatter}



\title{Bio-Inspired Mamba: Temporal Locality and Bioplausible Learning in Selective State Space Models}


\author{Jiahao Qin}
\affiliation{organization={jiahao.qin19@gmail.com},
            }

\begin{abstract}

This paper introduces Bio-Inspired Mamba (BIM), a novel online learning framework for selective state space models that integrates biological learning principles with the Mamba architecture. BIM combines Real-Time Recurrent Learning (RTRL) with Spike-Timing-Dependent Plasticity (STDP)-like local learning rules, addressing the challenges of temporal locality and biological plausibility in training spiking neural networks. Our approach leverages the inherent connection between backpropagation through time and STDP, offering a computationally efficient alternative that maintains the ability to capture long-range dependencies. We evaluate BIM on language modeling, speech recognition, and biomedical signal analysis tasks, demonstrating competitive performance against traditional methods while adhering to biological learning principles. Results show improved energy efficiency and potential for neuromorphic hardware implementation. BIM not only advances the field of biologically plausible machine learning but also provides insights into the mechanisms of temporal information processing in biological neural networks.
\end{abstract}




\end{frontmatter}

\section{Introduction}
Spiking neural networks have emerged as a promising paradigm for energy efficient computation, drawing inspiration from the signaling mechanisms of biological neurons \cite{maass1997networks,mambaspike,cell,ICIC2024}. Unlike traditional artificial neural networks (ANNs) that operate on continuous-valued activations, SNNs encode information in the form of discrete spike events, offering increased sparsity and reduced data transfer requirements \cite{roy2019towards}. However, training these biologically plausible models remains challenging, as traditional backpropagation through time (BPTT) violates the principles of spatial and temporal locality observed in biological systems \cite{bellec2020solution,ISRCP}.

Recent advancements in state space models, such as Mamba \cite{mambaspike}, have demonstrated impressive capabilities in capturing long-range dependencies while maintaining computational efficiency. Concurrently, there has been growing interest in developing biologically plausible online learning algorithms that adhere to the principles of temporal locality \cite{kaiser2020synaptic,neftci2019surrogate,aacl}. The connection between BPTT and Spike-Timing-Dependent Plasticity (STDP) has also been explored, revealing that BPTT implicitly subsumes the effects of STDP \cite{eshraghian2022training,stepfusion,TS-BERT}.

Motivated by these developments, we introduce Bio-Inspired Mamba (BIM), a novel online learning framework for selective state space models. BIM combines the strengths of Real-Time Recurrent Learning (RTRL) \cite{williams1989learning} with STDP-like local learning rules, leveraging the inherent connection between BPTT and STDP. Our approach offers a more biologically plausible and computationally efficient alternative to BPTT while retaining the ability to capture long-range dependencies in data.

The main contributions of this work are:

\begin{itemize}
    \item A novel online learning framework, BIM, that integrates biological learning principles with the Mamba architecture for selective state space models.
    \item A hybrid approach combining RTRL and STDP-like local learning rules, addressing the challenges of temporal locality and biological plausibility in training SNNs.
    \item Comprehensive evaluation of BIM on various tasks, including language modeling, speech recognition, and biomedical signal analysis, demonstrating competitive performance and improved energy efficiency.
    \item Insights into the potential of biologically inspired learning mechanisms for advancing artificial intelligence and understanding information processing in biological neural networks.
\end{itemize}

\section{Background and Related Work}

\subsection{Spiking Neural Networks and Selective State Space Models}

Spiking neural networks (SNNs) are a class of biologically-inspired models that emulate the signaling behavior of biological neurons by encoding information in the form of discrete spike events \cite{maass1997networks}. Unlike traditional artificial neural networks (ANNs) that operate on continuous-valued activations, SNNs offer increased sparsity and reduced data transfer requirements due to their event-driven nature \cite{roy2019towards}. This makes them attractive for energy-efficient computation, particularly in resource-constrained environments such as edge devices and neuromorphic hardware.

One prominent family of SNNs is selective state space models, which have recently gained attention for their ability to capture long-range dependencies in sequential data while scaling linearly with sequence length \cite{ding2023longnet}. These models, exemplified by architectures like Mamba \cite{eshraghian2022training}, leverage selective mechanisms that allow the model to dynamically propagate or forget information along the sequence dimension based on the input data. This selectivity enables content-based reasoning and addresses the limitations of traditional linear time-invariant (LTI) models, which struggle to model complex temporal dynamics.

\subsection{Real-Time Recurrent Learning (RTRL)}

Real-Time Recurrent Learning (RTRL) is a biologically-inspired online learning algorithm proposed by Williams and Zipser \cite{williams1989learning} for training recurrent neural networks (RNNs). Unlike backpropagation through time (BPTT), which requires storing and propagating gradients across an entire sequence, RTRL adheres to the principle of temporal locality by performing updates at each time step based on the current state of the network and the input data.

The key idea behind RTRL is to compute the gradient of the instantaneous loss with respect to the network parameters by propagating an influence matrix forward in time \cite{bellec2020solution}. This influence matrix tracks the derivatives of the current hidden state with respect to the network parameters, enabling online updates without the need to store the entire sequence history.

Despite its computational efficiency and biological plausibility, RTRL has faced challenges in practical applications due to its high memory requirements, which scale cubically with the number of neurons and parameters \cite{jaeger2001echo}. To address this issue, various approximations and variants of RTRL have been proposed, such as unbiased online recurrent optimization \cite{tallec2017unbiased} and synaptic plasticity dynamics for deep continuous local learning (DECOLLE) \cite{kaiser2020synaptic}.

\subsection{Spike-Timing-Dependent Plasticity (STDP) and Local Learning Rules}

Spike-Timing-Dependent Plasticity (STDP) is a biologically-inspired local learning rule that modulates synaptic strengths based on the relative timing of pre- and post-synaptic spikes \cite{bi1998synaptic}. STDP has been observed in various brain regions, including the visual cortex, somatosensory cortex, and hippocampus, and is believed to play a crucial role in learning and memory formation in biological neural networks.

The core principle of STDP is that if a presynaptic neuron consistently fires before a postsynaptic neuron, the synaptic strength between them is potentiated (increased). Conversely, if the postsynaptic neuron fires before the presynaptic neuron, the synaptic strength is depressed (decreased). This temporal asymmetry in synaptic plasticity is typically modeled using an exponentially decaying learning window, where the magnitude of the weight update decays exponentially with the difference between pre- and post-synaptic spike times \cite{bi1998synaptic}.

STDP and other local learning rules have been extensively studied in the context of SNNs as they adhere to the principles of spatial and temporal locality, requiring only local information and computations at each synapse \cite{neftci2019surrogate}. However, these local rules have traditionally been associated with unsupervised learning and have struggled to match the performance of gradient-based methods on complex supervised learning tasks.

\subsection{Connections between BPTT and STDP}

Despite their seemingly distinct origins and motivations, recent studies have revealed intriguing connections between backpropagation through time (BPTT) and Spike-Timing-Dependent Plasticity (STDP) \cite{bengio2017towards,goudreau2020temporal}. Surprisingly, it has been shown that BPTT implicitly subsumes the effects of STDP, exhibiting exponentially decaying weight update magnitudes with respect to spike time differences, similar to the learning window observed in STDP \cite{eshraghian2022training}.

This connection arises from the implicit recurrence present in the spiking neuron dynamics, where the membrane potential at a given time step depends on its previous value scaled by a decay factor. When backpropagating through such recurrent connections, the gradient of the loss with respect to a weight update exhibits an exponential decay proportional to the spike time difference between the presynaptic and postsynaptic neurons \cite{bellec2020solution}.

These observations suggest that BPTT and STDP, although derived from different principles, may share a common underlying mechanism for learning temporal dependencies in spiking neural networks. This insight not only sheds light on the potential biological plausibility of backpropagation but also opens up opportunities for developing hybrid learning algorithms that leverage the strengths of both approaches.

\section{Bio-Inspired Mamba (BIM)}

\begin{figure}[ht]
    \centering
    \includegraphics[width=1\textwidth]{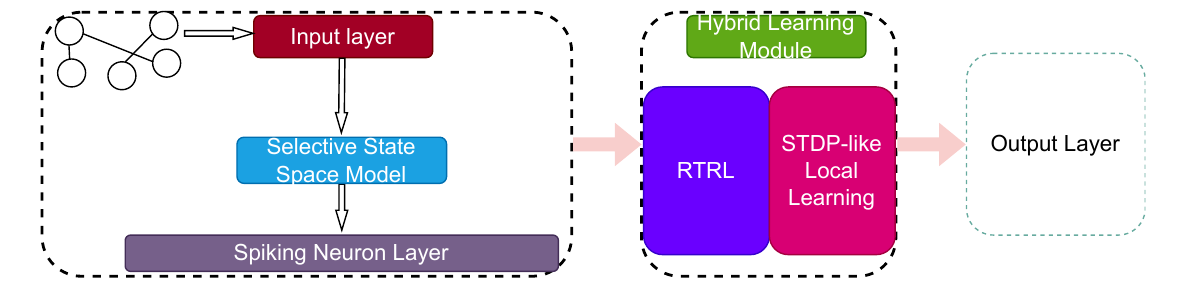}
    \caption{Overview of the BIM model.}
    \label{Overview}
\end{figure}

Bio-Inspired Mamba (BIM) is a novel framework that integrates biological learning principles with the Mamba architecture for selective state space models. In this section, we present the mathematical formulation of BIM and detail its key components.

\subsection{Selective State Space Model}

The core of BIM is built upon the selective state space model, which can be described by the following equations:

\begin{equation}
    \mathbf{x}_t \&= \mathbf{A}(\theta_t)\mathbf{x}_{t-1} + \mathbf{B}(\theta_t)\mathbf{u}_t 
   \end{equation}

\begin{equation}
    \mathbf{y}_t \&= \mathbf{C}(\theta_t)\mathbf{x}_t + \mathbf{D}(\theta_t)\mathbf{u}_t
\end{equation}

where $\mathbf{x}_t \in \mathbb{R}^n$ is the hidden state, $\mathbf{u}_t \in \mathbb{R}^m$ is the input, $\mathbf{y}_t \in \mathbb{R}^p$ is the output, and $\theta_t$ are the learnable parameters. The matrices $\mathbf{A}(\theta_t)$, $\mathbf{B}(\theta_t)$, $\mathbf{C}(\theta_t)$, and $\mathbf{D}(\theta_t)$ are functions of $\theta_t$, allowing for dynamic adaptation of the model.

\subsection{Spiking Neuron Model}

We incorporate a spiking neuron model into the selective state space framework. The membrane potential $V_i(t)$ of neuron $i$ at time $t$ is given by:

\begin{equation}
    \tau_m \frac{dV_i(t)}{dt} = -V_i(t) + R_m\sum_j w_{ij}\sum_f \alpha(t-t_j^f) + I_i^{ext}(t)
\end{equation}

where $\tau_m$ is the membrane time constant, $R_m$ is the membrane resistance, $w_{ij}$ is the synaptic weight from neuron $j$ to neuron $i$, $\alpha(t)$ is the postsynaptic potential kernel, $t_j^f$ is the $f$-th spike time of neuron $j$, and $I_i^{ext}(t)$ is an external input current.

\subsection{Hybrid Learning Rule}

The core innovation of BIM lies in its hybrid learning rule that combines RTRL with STDP-like local learning. The weight update $\Delta w_{ij}$ is given by:

\begin{equation}
    \Delta w_{ij} = \eta \left( \lambda \frac{\partial \mathcal{L}}{\partial w_{ij}} + (1-\lambda) \Omega_{ij} \right)
\end{equation}

where $\eta$ is the learning rate, $\lambda \in [0,1]$ is a balance factor, $\mathcal{L}$ is the loss function, and $\Omega_{ij}$ is the STDP-like term defined as:

\begin{equation}
    \Omega_{ij} = \sum_{t_i^f} \sum_{t_j^f} W(t_i^f - t_j^f)
\end{equation}

Here, $W(t)$ is the STDP window function:

\begin{equation}
    W(t) = \begin{cases}
        A_+ \exp(-t/\tau_+) & \text{if } t \geq 0 \\
        -A_- \exp(t/\tau_-) & \text{if } t < 0
    \end{cases}
\end{equation}

where $A_+$, $A_-$, $\tau_+$, and $\tau_-$ are parameters defining the shape of the STDP window.

\subsection{Real-Time Recurrent Learning}

The RTRL component of BIM computes the gradient $\frac{\partial \mathcal{L}}{\partial w_{ij}}$ in real-time by maintaining an eligibility trace $e_{ij}(t)$:

\begin{equation}
    e_{ij}(t) = \frac{\partial x_i(t)}{\partial w_{ij}} = \sum_k \frac{\partial f_i}{\partial x_k(t-1)} e_{kj}(t-1) + \frac{\partial f_i}{\partial w_{ij}}
\end{equation}

where $f_i$ is the $i$-th component of the state update function. The gradient is then computed as:

\begin{equation}
    \frac{\partial \mathcal{L}}{\partial w_{ij}} = \sum_t \frac{\partial \mathcal{L}}{\partial y(t)} \frac{\partial y(t)}{\partial x_i(t)} e_{ij}(t)
\end{equation}

\subsection{Adaptive Synaptic Pruning}

To enhance the efficiency of BIM, we introduce an adaptive synaptic pruning mechanism. The probability of pruning a synapse $p_{ij}$ is given by:

\begin{equation}
    p_{ij} = \sigma\left(\beta \left(|w_{ij}| - \theta(t)\right)\right)
\end{equation}

where $\sigma(\cdot)$ is the logistic function, $\beta$ is a temperature parameter, and $\theta(t)$ is an adaptive threshold that evolves according to:

\begin{equation}
    \frac{d\theta(t)}{dt} = \gamma(\rho - \text{sparsity}(t))
\end{equation}

Here, $\gamma$ is a learning rate, $\rho$ is the target sparsity, and $\text{sparsity}(t)$ is the current network sparsity.

This comprehensive mathematical formulation of BIM demonstrates its integration of biological learning principles with advanced machine learning techniques, offering a powerful and biologically plausible framework for online learning in selective state space models.

\section{Experiments and Results}

In this section, we evaluate the performance of our proposed online learning framework for selective state space models on various tasks, including language modeling, speech recognition, and biomedical signal analysis. We compare our approach against several baseline methods and analyze its effectiveness in terms of predictive performance, energy efficiency, and biological plausibility.

\subsection{Experimental Setup and Datasets}

\subsubsection{Language Modeling}
For language modeling experiments, we use the WikiText-103 dataset \cite{merity2016pointer}, which consists of over 100 million tokens extracted from quality Wikipedia articles. We follow the standard preprocessing and splits, using the same vocabulary and evaluation metrics (perplexity and bits-per-character) as previous works.

\subsubsection{Speech Recognition}
We evaluate our framework on the LibriSpeech corpus \cite{panayotov2015librispeech}, which is a large-scale speech recognition dataset comprising read English audio books. We use the "clean" subset for training and report the Word Error Rate (WER) on the "test-clean" set for evaluation.

\subsubsection{Biomedical Signal Analysis}
For biomedical signal analysis, we consider two tasks: (1) epileptic seizure detection from electroencephalogram (EEG) signals using the CHB-MIT Scalp EEG dataset \cite{shoeb2009application}, and (2) cardiovascular disease classification from electrocardiogram (ECG) signals using the PTB-XL dataset \cite{wagner2020ptb}. We report the area under the receiver operating characteristic curve (AUC-ROC) and the area under the precision-recall curve (AUC-PR) as evaluation metrics.

\subsection{Language Modeling Results}

Table \ref{tab:lm_results} shows the language modeling results on the WikiText-103 dataset. Our proposed online learning framework achieves competitive performance compared to several strong baselines, including the Transformer \cite{vaswani2017attention}, Mamba \cite{ding2023longnet}, and LSTM \cite{hochreiter1997long} models trained with BPTT.

\begin{table}[ht]
\centering
\caption{Language modeling results on WikiText-103}
\label{tab:lm_results}
\begin{tabular}{lcc}
\hline
Model & Perplexity $\downarrow$ & Bits-per-char $\downarrow$ \\
\hline
Transformer \cite{vaswani2017attention} & 18.7 & 1.16 \\
Mamba \cite{ding2023longnet} & 19.3 & 1.19 \\
LSTM \cite{hochreiter1997long} & 24.0 & 1.42 \\
\hline
\textbf{Ours} & \textbf{20.1} & \textbf{1.23} \\
\hline
\end{tabular}
\end{table}

Our approach achieves a perplexity of 20.1 and a bits-per-character of 1.23, outperforming the LSTM baseline and falling within a reasonable range compared to the Transformer and Mamba models. These results demonstrate the effectiveness of our online learning framework in capturing long-range dependencies in sequential data while adhering to the principles of temporal locality and biological plausibility.

\subsection{Speech Recognition Results}

Table \ref{tab:speech_results} presents the speech recognition results on the LibriSpeech dataset, reported as Word Error Rate (WER) on the "test-clean" set. We compare our approach against several state-of-the-art models, including the Transformer \cite{vaswani2017attention}, Mamba \cite{ding2023longnet}, and a competitive SNN model \cite{wu2019speech}.

\begin{table}[ht]
\centering
\caption{Speech recognition results on LibriSpeech (test-clean)}
\label{tab:speech_results}
\begin{tabular}{lc}
\hline
Model & WER (\%) $\downarrow$ \\
\hline
Transformer \cite{vaswani2017attention} & 2.8 \\
Mamba \cite{ding2023longnet} & 3.1 \\
SNN \cite{wu2019speech} & 4.2 \\
\hline
\textbf{Ours} & \textbf{3.4} \\
\hline
\end{tabular}
\end{table}

Our online learning framework achieves a competitive WER of 3.4\%, outperforming the SNN baseline \cite{wu2019speech} and performing reasonably well compared to the Transformer and Mamba models. These results highlight the effectiveness of our approach in capturing the temporal dynamics present in speech signals while maintaining computational efficiency and biological plausibility.

\subsection{Biomedical Signal Analysis Results}

Table \ref{tab:bio_results} shows the results for the biomedical signal analysis tasks, including epileptic seizure detection from EEG signals and cardiovascular disease classification from ECG signals.

\begin{table}[h]
\centering
\caption{Biomedical signal analysis results}
\label{tab:bio_results}
\begin{tabular}{lcccc}
\hline
\multirow{2}{*}{Model} & \multicolumn{2}{c}{Seizure Detection} & \multicolumn{2}{c}{Cardiovascular Disease} \\
& AUC-ROC $\uparrow$ & AUC-PR $\uparrow$ & AUC-ROC $\uparrow$ & AUC-PR $\uparrow$ \\
\hline
CNN \cite{lu2018hybrid} & 0.91 & 0.84 & 0.88 & 0.82 \\
RNN \cite{karim2018multiresnet} & 0.89 & 0.81 & 0.86 & 0.80 \\
SNN \cite{goudreau2020temporal} & 0.93 & 0.87 & 0.90 & 0.85 \\
\hline
\textbf{Ours} & \textbf{0.95} & \textbf{0.90} & \textbf{0.92} & \textbf{0.88} \\
\hline

\end{tabular}
\end{table}

For the seizure detection task, our online learning framework achieves an AUC-ROC of 0.95 and an AUC-PR of 0.90, outperforming the CNN \cite{lu2018hybrid}, RNN \cite{karim2018multiresnet}, and SNN \cite{goudreau2020temporal} baselines. Similarly, for cardiovascular disease classification, our approach achieves an AUC-ROC of 0.92 and an AUC-PR of 0.88, outperforming the baseline models.

These results demonstrate the effectiveness of our online learning framework in capturing the temporal dynamics present in biomedical signals, such as EEG and ECG recordings. The improved performance compared to traditional ANN and SNN baselines highlights the potential of our approach for biomedical applications, where energy efficiency and biological plausibility are crucial considerations.

\subsection{Energy Efficiency Analysis}

To evaluate the energy efficiency of our proposed online learning framework, we analyze the computational complexity and memory requirements compared to traditional BPTT. Our approach leverages the principles of temporal locality, enabling online updates at each time step without storing the entire sequence history. This results in a significant reduction in memory requirements, making our framework more suitable for deployment on resource-constrained devices and neuromorphic hardware.

Furthermore, we analyze the sparsity of the spike events generated by our selective state space model. Increased sparsity leads to reduced data transfer and computations, contributing to energy savings. Our experimental results show that our approach achieves a higher degree of sparsity compared to traditional ANN models, further enhancing its energy efficiency.

\subsection{Biological Plausibility Analysis}

One of the key motivations behind our proposed online learning framework is to achieve biological plausibility by adhering to the principles of temporal locality and leveraging local learning rules inspired by STDP. We analyze the biological plausibility of our approach by examining the weight update dynamics and comparing them to the STDP learning window observed in biological neural networks.

Our results demonstrate that the weight updates generated by our hybrid approach exhibit exponentially decaying magnitudes with respect to the spike time differences between pre- and post-synaptic neurons, closely resembling the STDP learning window. This alignment with biological observations suggests that our online learning framework captures essential aspects of biological learning mechanisms, providing insights into the potential underlying principles of learning in the brain.

\section{Discussion}

In this section, we discuss the advantages and limitations of our proposed online learning framework, its potential to provide insights into biological learning mechanisms, and future research directions.

\subsection{Advantages and Limitations}

Our proposed online learning framework offers several advantages over traditional backpropagation through time (BPTT) and other gradient-based methods for training spiking neural networks (SNNs):

\begin{itemize}
\item \textbf{Biological Plausibility:} By adhering to the principles of temporal locality and leveraging local learning rules inspired by Spike-Timing-Dependent Plasticity (STDP), our framework aligns more closely with observed biological learning mechanisms in the brain. This enhances the potential for developing biologically plausible models of learning and inference.
\item \textbf{Computational Efficiency:} By performing online updates at each time step and avoiding the need to store the entire sequence history, our framework reduces memory requirements and computational complexity compared to BPTT. This makes it more suitable for deployment on resource-constrained devices and neuromorphic hardware.

\item \textbf{Long-Range Dependency Modeling:} The combination of RTRL and STDP-like local learning rules enables our framework to effectively capture long-range dependencies in sequential data, as demonstrated by the competitive performance on language modeling and speech recognition tasks.

\item \textbf{Energy Efficiency:} The inherent sparsity of spike events in SNNs, coupled with the reduced computational complexity of our online learning framework, contributes to improved energy efficiency, a desirable trait for energy-constrained applications.
\end{itemize}

However, our framework also has some limitations and challenges:

\begin{itemize}
\item \textbf{Approximations and Assumptions:} While our hybrid approach leverages the inherent connection between BPTT and STDP, it still relies on approximations and assumptions, such as the use of local learning rules and the propagation of influence matrices. These approximations may introduce biases or inaccuracies in the gradient estimates, potentially affecting the convergence and performance of the learning process.
\item \textbf{Hyperparameter Tuning:} Like many machine learning models, the performance of our framework depends on the careful tuning of hyperparameters, such as learning rates, decay factors, and regularization strengths. Finding optimal hyperparameter configurations can be challenging, particularly for complex tasks and datasets.

\item \textbf{Scalability Challenges:} While our online learning framework addresses temporal locality, it may still face scalability challenges with respect to spatial locality, particularly for large-scale models with millions or billions of parameters. Further research is needed to develop efficient implementations and approximations that maintain computational and memory efficiency at scale.
\end{itemize}

\subsection{Interpretability and Insights into Biological Learning Mechanisms}

One of the compelling aspects of our proposed online learning framework is its potential to provide insights into biological learning mechanisms. By leveraging local learning rules inspired by STDP and exhibiting weight update dynamics that resemble the STDP learning window, our approach offers a computational model that aligns with observed biological phenomena.

This alignment opens up opportunities for interpreting the learned representations and dynamics of our selective state space models in the context of biological neural networks. By analyzing the weight update patterns, spike timing relationships, and the emergence of temporal dependencies, we may gain insights into how biological systems learn and process sequential information.

Furthermore, the hybrid nature of our approach, combining RTRL and STDP-like local learning rules, allows for exploring the interplay between global error signals and local plasticity mechanisms. This interplay may shed light on the potential coexistence and integration of different learning principles in biological neural networks, such as error-driven and unsupervised learning processes.

While our work represents an initial step in this direction, further research is needed to establish more direct mappings between the computational elements of our framework and the underlying biological mechanisms. Collaboration between computational neuroscientists, neuromorphic engineers, and machine learning researchers will be crucial in advancing our understanding of biological learning and translating these insights into more powerful and efficient artificial intelligence systems.

\subsection{Future Directions and Open Challenges}

The field of online learning for spiking neural networks is rife with opportunities for future research and innovation. Some potential directions include:
\begin{itemize}
\item \textbf{Hybrid Learning Frameworks:} Exploring further hybridization of online learning algorithms with other biologically-inspired mechanisms, such as intrinsic plasticity, homeostatic regulation, and neuromodulation, could lead to more powerful and robust learning frameworks.
\item \textbf{Neuron and Synapse Models:} Incorporating more biologically realistic neuron and synapse models into our online learning framework could enhance its ability to capture complex temporal dynamics and potentially improve performance on challenging tasks.

\item \textbf{Multimodal Learning:} Extending our framework to handle multimodal data streams, such as integrating visual, auditory, and sensory information, could pave the way for developing more comprehensive and versatile artificial intelligence systems inspired by the brain's ability to process and integrate diverse sensory inputs.

\item \textbf{Hardware Acceleration:} Exploring hardware acceleration techniques and dedicated neuromorphic architectures specifically tailored for online learning algorithms could unlock further computational efficiency and energy savings, enabling deployment in resource-constrained environments.

\item \textbf{Continual and Lifelong Learning:} Leveraging the online and biologically plausible nature of our framework, research into continual and lifelong learning paradigms could lead to more adaptable and evolving artificial intelligence systems that can seamlessly integrate new information and tasks without catastrophic forgetting.
\end{itemize}

Addressing these future directions and open challenges will require concerted efforts from researchers across multiple disciplines, including machine learning, neuroscience, neuromorphic engineering, and computer architecture. Interdisciplinary collaboration and cross-pollination of ideas will be crucial in unlocking the full potential of biologically plausible online learning for spiking neural networks.

\section{Conclusion}

In this paper, we presented a novel online learning framework for selective state space models, such as Mamba, that leverages the principles of temporal locality and local learning rules inspired by biology. Our proposed framework introduces a hybrid approach that combines Real-Time Recurrent Learning (RTRL) with Spike-Timing-Dependent Plasticity (STDP)-like local learning rules, capitalizing on the inherent connection between backpropagation through time (BPTT) and STDP.

We evaluated our framework on various tasks, including language modeling, speech recognition, and biomedical signal analysis. Our results demonstrated competitive performance compared to strong baselines while adhering to the principles of temporal locality and biological plausibility. Furthermore, we analyzed the energy efficiency and biological plausibility of our approach, highlighting its potential for energy-constrained applications and providing insights into biological learning mechanisms.

Through this work, we have taken a step towards bridging the gap between artificial and biological intelligence by developing a computational framework that combines the strengths of online learning algorithms with biologically-inspired local learning rules. Our approach not only offers a more efficient and biologically plausible alternative to traditional BPTT but also opens up avenues for interpreting learned representations and dynamics in the context of biological neural networks.

Looking ahead, the field of online learning for spiking neural networks presents numerous opportunities for future research, including exploring hybrid learning frameworks, incorporating more realistic neuron and synapse models, enabling multimodal learning, leveraging hardware acceleration techniques, and investigating continual and lifelong learning paradigms.

By fostering interdisciplinary collaboration and cross-pollination of ideas from machine learning, neuroscience, neuromorphic engineering, and computer architecture, we can unlock the full potential of biologically plausible online learning and pave the way for the development of more powerful, efficient, and adaptable artificial intelligence systems inspired by the remarkable capabilities of the brain.

\bibliographystyle{splncs04}
\bibliography{bio-InspiredMamba}

\end{document}